# Multilinear Biased Discriminant Analysis: A Novel Method for Facial Action Unit Representation

Mahmoud Khademi †, Mehran Safayani † and Mohammad T. Manzuri-Shalmani †

† : Sharif University of Tech., DSP Lab, `{khademi@ce, safayani@ce, manzuri@}.sharif.edu`

**Abstract** In this paper a novel efficient method for representation of facial action units by encoding an image sequence as a fourth-order tensor is presented. The multilinear tensor-based extension of the biased discriminant analysis (BDA) algorithm, called multilinear biased discriminant analysis (MBDA), is first proposed. Then, we apply the MBDA and two-dimensional BDA (2DBDA) algorithms, as the dimensionality reduction techniques, to Gabor representations and the geometric features of the input image sequence respectively. The proposed scheme can deal with the asymmetry between positive and negative samples as well as curse of dimensionality dilemma. Extensive experiments on Cohn-Kanade database show the superiority of the proposed method for representation of the subtle changes and the temporal information involved in formation of the facial expressions. As an accurate tool, this representation can be applied to many areas such as recognition of spontaneous and deliberate facial expressions, multi modal/media human computer interaction and lie detection efforts.

## 1 Introduction

Human face-to-face communication is a standard of perfection for developing a natural, robust, effective and flexible multi modal/media human-computer interface due to multimodality and multiplicity of its communication channels. In this type of communication, the facial expressions constitute the main modality [1]. In this regard, automatic facial expression analysis can use the facial signals as a new modality and causes the interaction between human and computer more robust and flexible. Moreover, it can be used in other areas such as lie detection, neurology and clinical psychology.

Facial expression analysis includes both measurement of facial motion (e.g. mouth stretch or outer brow raiser) and recognition of expression (e.g. surprise or anger). Real-time fully automatic facial expression analysis is a challenging complex topic in computer vision due to pose variations, illumination variations, different age, gender, ethnicity, facial hair, occlusion, head motions and lower intensity of expressions. Two survey papers summarized the work of facial expression analysis [2, 3]. Regardless of the face detection stage, a typical automatic facial expression analysis consists of facial expression data extraction and facial expression classification stages. Facial feature processing may happen either holistically, where the face is processed as a whole, or locally. Holistic feature extraction methods are good at determining prevalent facial expressions, whereas local methods are able to detect subtle changes in small areas.

There are mainly two methods for facial data extraction: geometric feature-based methods and appearance-based method. The geometric facial features present the shape and locations of facial components. With appearance-based methods, image filters, e.g. Gabor wavelets, are applied to either the whole face or specific regions in a face image to extract a feature vector [4]. The sequence-based recognition method uses the temporal information of the sequences to recognize the expressions of one or more frames. To use the temporal information, the techniques such as hidden Markov models (HMMs), recurrent neural networks and rule-based classifier were applied.

The main goal of this paper is developing an accurate method for representation of facial action units (AUs). Our method has the following characteristics: 1) it can deal with the asymmetry between positive and negative image sequences as well as curse of dimensionality dilemma, 2) multiple interrelated subspaces can cooperate to discriminate positive and negative image sequences in the new subspace, 3) increasing the recognition rate by using both geometric and appearance features, and 4) it is robust to illumination changes and can represent subtle changes in facial muscles as well as temporal information involved in formation of the expressions.

The rest of the paper has been organized as follows: In section 2, we review the related works. In section 3, we first describe the proposed approach for representation of facial action units using fourth-order tensors. Then, multilinear biased discriminant analysis (MBDA) algorithm, for reducing the dimensionality of the tensor in all directions, is proposed. Section 4 reports our experimental results, and section 5 presents conclusions and a discussion.

## 2 Related Works

2.1 Facial Action Coding System

The facial action coding system (FACS) is a system developed by Ekman and Friesen to detect subtle changes in facial features. The FACS is composed of 44 facial action units (AUs). 30 AUs of them are related to movement of a specific set of facial muscles: 12 for upper face (e.g. AU 1 inner brow raiser, AU 2 outer brow raiser, AU 4 brow lowerer, AU 5 upper lid raiser, AU 6 cheek raiser, AU 7 lid tightener) and 18 for lower face (e.g. AU 9 nose wrinkle, AU 10 upper lip raiser, AU 12 lip corner puller, AU 15 lip corner depressor, AU 17 chin raiser, AU 20 lip stretcher, AU 25 lips part, AU 26 jaw drop, AU 27 mouth stretch). Facial action units can occur in a combinational manner and vary in intensity. Although the number of atomic action units is relatively small, more than 7000 different AU combinations have been observed (for more details see [5]).

2.2 Biased Learning

Biased learning is a learning problem in which there are an unknown number of classes but we are only interested in one class which is called "positive" class. Other samples are considered as "negative" samples. In fact, these samples can come from an uncertain number of classes. Suppose $\{x_i | i = 1, \ldots, N_x\}$ and $\{y_i | i = 1, \ldots, N_y\}$ are the set of positive and negative $d$-dimensional samples (feature vectors) respectively. Consider the problem of finding $d \times r$ transformation matrix $w$ ($r \ll d$), such that separates projected positive samples from projected negatives in the new subspace.

The dimension reduction methods like fisher discriminant analysis (FDA) and multiple discriminant analysis have addressed this problem simply as a two-class classification problem with symmetric treatment on positive and negative examples. For example in FDA, the goal is to find a subspace in which the ratio of between-class scatter over within-class scatter matrices is maximized. However, it is part of the objective function that negative samples shall cluster in the discriminative subspace. This is an unnecessary and potentially damaging requirement because very likely the negative samples belong to multiple classes and any constraint put on them other than stay away from the positives is unnecessary and misleading. With asymmetric treatment toward the positive samples, Zhou and Huang [6] proposed the following objective function:

$$w_{opt} = \arg\max_W \frac{\text{trace}(w^T S_y w)}{\text{trace}(w^T S_x w)} \quad (1)$$

where $S_y$ and $S_x$ are within-class scatter matrices of negative and positive samples with respect to positive centroid, respectively. The goal is to find $w$ that clusters only positive samples while keeping negatives away.

Our goal is developing a facial action unit recognition system[1] that can detect whether the AUs occur or not. The input of the system is a sequence of frames from natural face towards one of the facial expressions with maximum intensity. Suppose we have extracted a feature matrix or a feature vector from each frame. In order to embed facial features in a low-dimensionality space and deal with curse of dimensionality problem, we should use a dimension reduction method. For recognition of each AU, we are facing an asymmetric two-class classification problem. For example when the goal is detecting whether AU 27 (mouth stretch) occur or not, the positive class includes all of sequences in the train set that represent stretching of the mouth; other sequences are considered as negative samples. These samples can come from an uncertain number of classes. They can represent any AU or AU combinations except AU 27. In fact, our problem is a biased learning problem. However, for applying the Zhou and Huang's method to the facial action unit recognition problem we should first transform the feature matrices or a feature vectors of the input image sequence into a one-dimensional vector that ignores the underlying data structure (temporal and local information) and leads to the curse of dimensionality dilemma and the small sample size problem.

## 3 The proposed Method

3.1 Representation of the Facial Action Units Using Fourth-Order Tensors

We represent an image sequence by a fourth-order tensor. The size of the cropped images is $m \times n$. $m$ and $n$ are length and width of the AU images in Fig 1. $t$ is number of the frames. The first frame represents without expression cropped image and the last frame represents the cropped image with maximum intensity of the expression. This direction of the tensor represents temporal information. In order to extract the facial features from each frame, we use a set of Gabor wavelets [4]. They allow detecting line endings and edge borders of each frame over multiple scales and with different orientations. Gabor wavelets also remove most of the variability in images that occur due to lighting changes.

---

[1] In some literatures "facial action unit detection" is used instead of "facial action unit recognition".

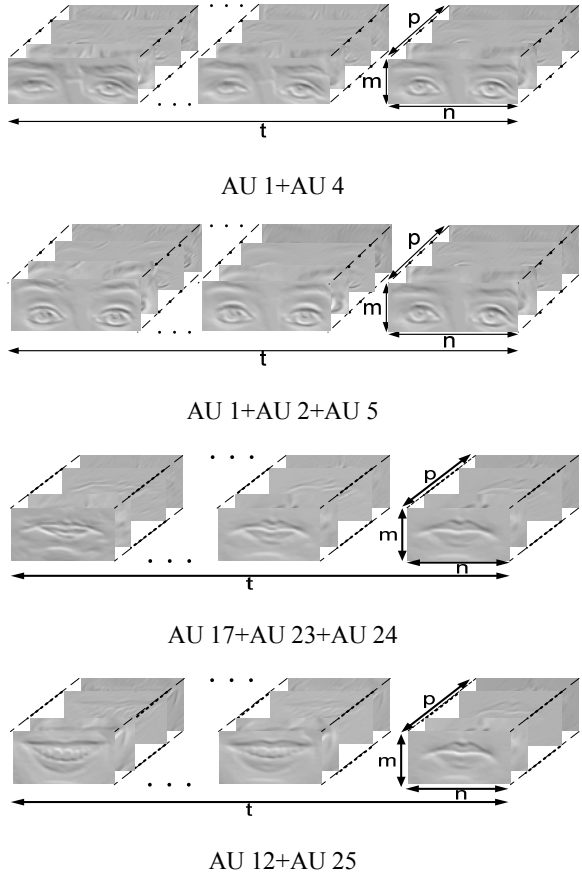

**Fig. 1.** Representation of the image sequences using fourth-order tensor and Gabor wavelets

Before explaining the MBDA algorithm, we define some notations; we represent tensors and matrices by uppercase and normal symbols respectively. $\text{unf}^{(j)}(\mathbf{Z})$ is j-mode unfolded matrix from tensor $\mathbf{Z}$. Fig. 2 shows a fourth-order tensor and its unfolded matrices. For a fourth-order tensor $\mathbf{Z}$ of dimension $d_1 \times d_2 \times d_3 \times d_4$, $\text{unf}^{(j)}(\mathbf{Z})$ is a $d_j \times (d_1)\dots(d_{j-1})(d_{j+1})\dots(d_4)$ matrix and $\text{unf}^{(j,k)}(\mathbf{Z})$ is the kth column of this matrix. $\text{col}(w)$ is number of columns of matrix w. $\text{fol}^{(j)}(w)$ is a tensor which is constructed by j-mode folding the columns of the matrix w. In fact, $\text{fol}^{(j)}\left(\text{unf}^{(j)}(\mathbf{Z})\right) = \mathbf{Z}$ for $j = 1,\dots,4$. The j-mode product of a tensor $\mathbf{Z}$ and a matrix w, is defined as:

$$\mathbf{Z} \times^j w = \text{fol}^{(j)}\left(w \times \text{unf}^{(j)}(\mathbf{Z})\right) \quad j = 1,\dots,4 \quad (2)$$

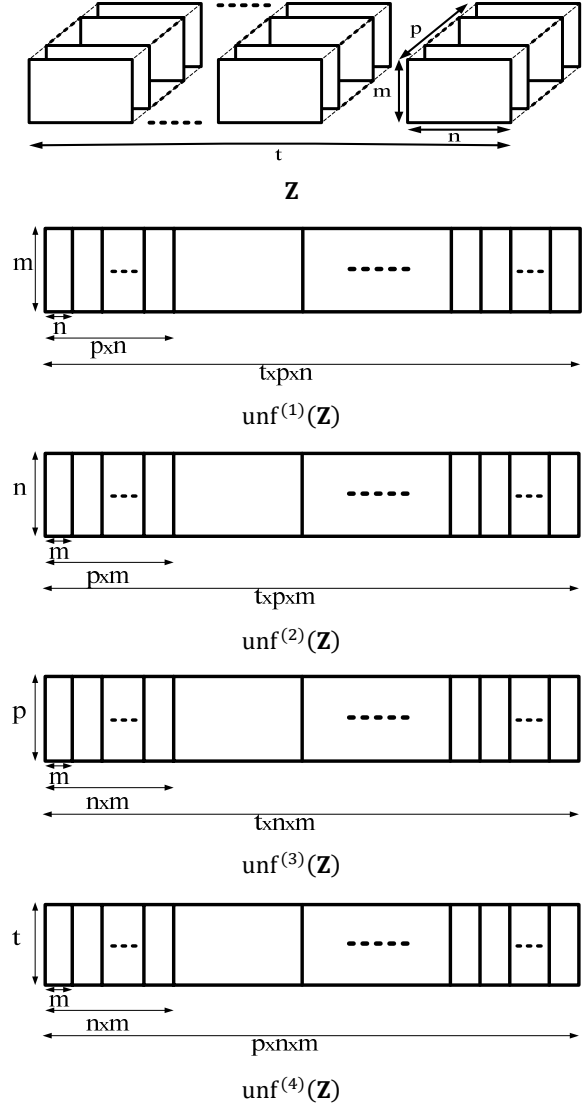

**Fig. 2.** A fourth-order tensor and its unfolded matrices

Finally, the square of the distance between two tensors $\mathbf{Z}_1$ and $\mathbf{Z}_2$ is as follows:

$$||\mathbf{Z}_1 - \mathbf{Z}_2||^2 = \sum_{i=1,j=1,k=1,s=1}^{d_1,d_2,d_3,d_4} \left(\mathbf{Z}_{1_{i,j,k,s}} - \mathbf{Z}_{2_{i,j,k,s}}\right)^2 \quad (3)$$

### 3.2 Multilinear Biased Discriminant Analysis (MBDA)

In this section we propose an extension of the objective function described by (1). Then, we discuss an iterative algorithm to find a solution for the nonlinear optimization problem described by the extended objective function. In order to develop a subject-independent facial action unit recognition system, we use difference Gabor responses instead of the Gabor responses. Difference Gabor responses are obtained by subtracting the Gabor representation of the natural expression face. This preprocessing stage changes the dimension of the tensor in temporal direction to $t - 1$. The output of the algorithm is four matrices (discriminative subspaces) which are used for reducing the dimensionality of the tensors in all directions.

Suppose $\{\mathbf{X}_i | i = 1, \ldots, N_x\}$ and $\{\mathbf{Y}_i | i = 1, \ldots, N_y\}$ are the set of positive and negative image sequences respectively. The following objective function is designed to seek four matrices (discriminative subspaces) which minimize the scatter of positive samples while maximize the scatter of negatives with respect to positive centroid $\mathbf{M}_x$:

$$\langle w_1, \ldots, w_4 \rangle_{opt} = \arg\max_{\langle w_1, \ldots, w_4 \rangle}$$

$$\frac{\sum_{i=1}^{N_y} ||\mathbf{Y}_i \times^1 w_1 \cdots \times^4 w_4 - \mathbf{M}_x \times^1 w_1 \cdots \times^4 w_4||^2}{\sum_{i=1}^{N_x} ||\mathbf{X}_i \times^1 w_1 \cdots \times^4 w_4 - \mathbf{M}_x \times^1 w_1 \cdots \times^4 w_4||^2} \quad (4)$$

This nonlinear optimization problem has no closed-form solution. We adopt an iterative approach to find the matrices. Each iteration of the algorithm composed of four optimization problems, i.e. j-mode optimization problems for $j = 1, \ldots 4$. Suppose $w_1, \ldots, w_{j-1}, w_{j+1}, \ldots, w_4$ are fixed. Define:

$$\mathbf{X}'_i = \mathbf{X}_i \times^1 w_1 \cdots \times^{j-1} w_{j-1} \times^{j+1} w_{j+1} \cdots \times^4 w_4 \quad (5)$$

The definition of $\mathbf{M}'_x$ and $\mathbf{Y}'_i$ are similar. The j-mode optimization problem is as follows:

$$\langle w_j \rangle_{opt} = \arg\max_{\langle w_j \rangle}$$

$$\frac{\sum_{i=1}^{N_y} ||\mathbf{Y}'_i \times^j w_j - \mathbf{M}'_x \times^j w_j||^2}{\sum_{i=1}^{N_x} ||\mathbf{X}'_i \times^j w_j - \mathbf{M}'_x \times^j w_j||^2} \quad j = 1, \ldots, 4 \quad (6)$$

The initial value of $w_1, w_2, w_3$ and $w_4$ are $n \times n, m \times m, p \times p$ and $t \times t$ identity matrices respectively. After solving each j-mode optimization problem the value of $w_j$ is updated immediately.

**Theorem.** The j-mode optimization problem, i.e. (6), can be rewritten as:

$$\langle w_j \rangle_{opt} = \arg\max_{\langle w_j \rangle}$$

$$\frac{\text{trace}(w_j^T S_y^{(j)} w_j)}{\text{trace}(w_j^T S_x^{(j)} w_j)} \quad j = 1, \ldots, 4 \quad (7)$$

where

$$S_y^{(j)} = \sum_{i=1}^{N_y} \sum_{k=1}^{\text{col}(\text{unf}^{(j)}(\mathbf{Y}'_i))} (\text{unf}^{(j,k)}(\mathbf{Y}'_i) - \text{unf}^{(j,k)}(\mathbf{M}'_x))(\text{unf}^{(j,k)}(\mathbf{Y}'_i) - \text{unf}^{(j,k)}(\mathbf{M}'_x))^T \quad (8)$$

$$S_x^{(j)} = \sum_{i=1}^{N_x} \sum_{k=1}^{\text{col}(\text{unf}^{(j)}(\mathbf{X}'_i))} (\text{unf}^{(j,k)}(\mathbf{X}'_i) - \text{unf}^{(j,k)}(\mathbf{M}'_x))(\text{unf}^{(j,k)}(\mathbf{X}'_i) - \text{unf}^{(j,k)}(\mathbf{M}'_x))^T \quad (9)$$

This can be achieved by (2), (3) and some algebraic manipulations. According to this theorem, the problem of finding optimal $w_j$ in (6) becomes finding the generalized eigenvectors α's associated with the largest eigenvalues λ's in the below generalized eigenanalysis problem:

$$S_y^{(j)} \alpha = \lambda S_x^{(j)} \alpha \quad (10)$$

The α's are columns of $w_j$ and the number of the selected α's in the first iteration, determines the low-dimensional representation of the image sequences in jth direction of the tensor. In other words, if we select $d_j$ eigenvectors for jth optimization problem, then the dimension of $w_1, w_2, w_3$ and $w_4$ would be $n \times d_1, m \times d_2, p \times d_3$ and $t \times d_4$ after the first iteration. To avoid singularity in the matrices and tuning down the influence of negative samples, the regularization and discounting may be required. Moreover, the weighting of the eigenvectors by the square roots of corresponding eigenvalues has good properties (for details see [7]). Although (1) and (7) are similar, the representation of the image sequences and metric are different.

## 4 Experimental Results

To evaluate the performance of the proposed method (MBDA) and other methods like two-dimensional biased discriminant analysis (2DBDA) [7] and multilinear discriminant analysis (MDA) [8], we tested them on Cohn-Kanade database [9]. The database includes 490 frontal view image sequences from over 97 subjects. Of theses, 300 sequences were used as the training set. Also, for upper face and lower face AUs, 120 and 140 sequences were used as the test set respectively. None of the test subjects appeared in training data set. Some of the sequences contained limited head motion. The final frame of each image sequence was coded using Facial Action Coding System which describes subject's expression in terms of action units. Image sequences from neutral to the frame with maximum intensity, were cropped into $57 \times 102$ and $52 \times 157$ pixel arrays for lower face and upper face action units respectively. Three intermediate frames, plus the first and last frames, were used. To extract appearance features we used 16 Gabor kernels. As a result, the tensor for representing the input sequence in the MBDA and MDA methods was of dimension $57 \times 102 \times 16 \times 4$ and $52 \times 157 \times 16 \times 4$ for lower face and upper face action units respectively (by considering the difference Gabor responses). In these two methods low-dimensional representation for a test data $\mathbf{Z}$ was derived as:

$$\tilde{\mathbf{Z}} = \mathbf{Z} \times^1 w_1 \times^2 w_2 \times^3 w_3 \times^4 w_4 \quad (11)$$

The output tensors were of dimension $3 \times 4 \times 1 \times 1$. The vectorized representation of $\tilde{\mathbf{Z}}$ was used as the appearance-based representation of the input sequence.

In order to extract geometric features we used a facial feature extraction method presented in [10]. The points of a 113-point grid which is called Wincanide-3, were placed on the first frame manually. Automatic registering of the grid with the face has been addressed in many literatures (e.g. see [11]). For upper face and lower face action units a particular set of points were selected. The pyramidal optical flow tracker [12] was employed to track the points of the model in the successive frames towards the last frame (see Fig. 3).

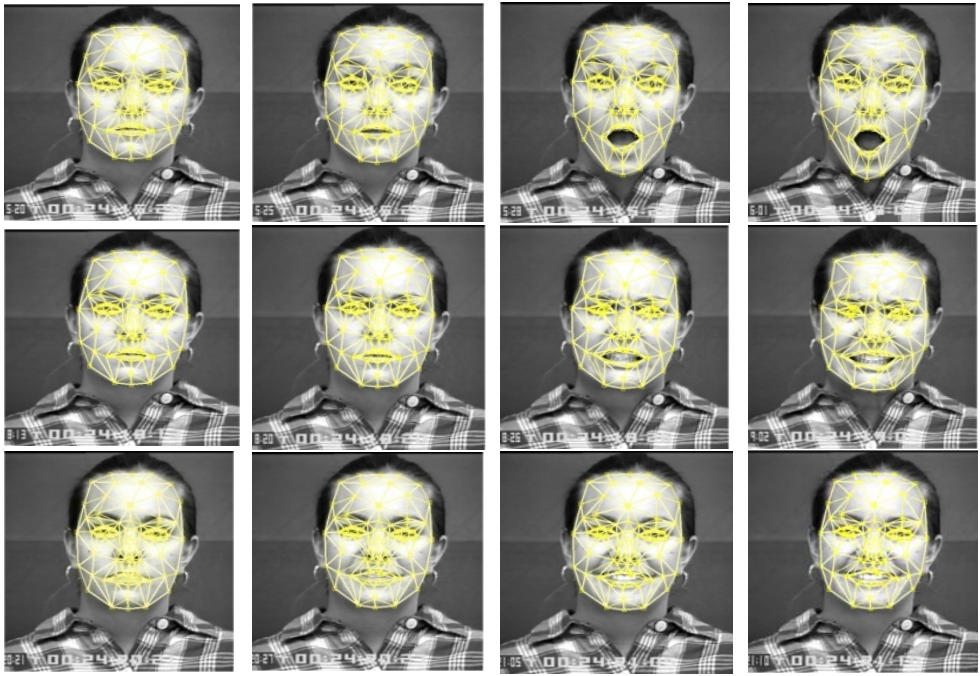

**Fig. 3.** Geometric-based facial feature extraction using grid tracking

The loss of the tracked points was handled through a model deformation procedure (for detail see [10]). For each frame the displacements of the points, with respect to the first frame, in two directions were calculated and placed in the columns of a matrix (each column for a frame). Then, we applied 2DBDA algorithm [7] to this matrix in two directions. The output matrix of 2DBDA algorithm was vectorized and concatenated to the feature vector which is resulted from appearance-based feature extraction method. A support vector machines (SVMs) classifier with Gaussian kernel was used to classify the samples in all experiments.

Table 1 and Table 2 show the upper face and lower face action unit recognition results respectively. We used geometric features in all methods. In 2DBDA+BDA method, we first applied 2DBDA algorithm [7] to each Gabor response of each frame separately in two directions. Then, we concatenated the vectorized representation of feature matrices of the sequence, and applied BDA algorithm to resulted vector. Although this method adopts an asymmetric method to separate positive samples from negatives in the new subspace, it ignores the
5

underlying structure (especially temporal information). Also, this method suffers from curse of dimensionality dilemma due to use of one-dimensional BDA algorithm.

In MDA method, we apply MDA algorithm [8] to forth-order tensors. Although this method keeps the underling structure, like FDA it addresses the problem simply as a two-class classification problem with symmetric treatment on positive and negative sequences. Moreover, in order to reveal the role of the appearance features, we did an experiment using geometric-based features only. In the proposed method (MBDA), an average recognition rate of 89.2 and 96.4 percent were achieved for upper face and lower face action units respectively. Also, an average false alarm rate of 6.7 and 2.1 percent were achieved for upper face and lower face action units respectively.

## 5 Conclusion and Discussion

We proposed an efficient method for representation of the facial AUs using a novel dimension reduction method. Although the computational cost of the proposed method can be high in training phase, it needs only four matrix products to reduce the dimensionality of the tensor in the test phase (see equation (11)). Employing a 3× 3 Gabor kernel and a grid with low number of vertices, we can construct the tensor of the input sequence and track the grid in less than two seconds, with moderate computing power. As a result, the proposed system is suitable for real-time applications. Future research direction is to consider variations on face pose in the tracking algorithm.

**Table 1.** Upper face action unit recognition results (R= recognition rate, F= false alarm)

| AUs | Sequences | Appearance (MBDA)+ Geometric (2DBDA) | | |
|---|---|---|---|---|
| | | True | Missing or extra | False |
| 1 | 10 | 8 | 1(1+2+4), 1(1+2) | 0 |
| 2 | 5 | 3 | 1(1+2+4), 1(1+2) | 0 |
| 4 | 10 | 10 | 0 | 0 |
| 5 | 10 | 10 | 0 | 0 |
| 6 | 10 | 10 | 0 | 0 |
| 7 | 5 | 4 | 0 | 1(6) |
| 1+2 | 20 | 19 | 1(2) | 0 |
| 1+2+4 | 10 | 10 | 0 | 0 |
| 1+2+5 | 5 | 4 | 1(1+2) | 0 |
| 1+4 | 5 | 4 | 1(1+2+4) | 0 |
| 1+6 | 5 | 4 | 1(1+6+7) | 0 |
| 4+5 | 10 | 9 | 1(4) | 0 |
| 6+7 | 15 | 12 | 1(1+6+7), 2(7) | 0 |
| Total | 120 | 107 | 12 | 1 |
| R | | 89.2% | | |
| F | | 6.7% | | |

| AUs | Sequences | Appearance (2DBDA [7]+BDA [6])+ Geometric (2DBDA) | | |
|---|---|---|---|---|
| | | True | Missing or extra | False |
| 1 | 10 | 6 | 2(1+2+4), 1(1+2) | 1(2) |
| 2 | 5 | 3 | 1(1+2+4), 1(1+2) | 0 |
| 4 | 10 | 9 | 1(1+2+4) | 0 |
| 5 | 10 | 10 | 0 | 0 |
| 6 | 10 | 8 | 1(1+6) | 1(7) |
| 7 | 5 | 3 | 1(6+7) | 1(6) |
| 1+2 | 20 | 17 | 1(2), 1(1+2+4) | 1(4) |
| 1+2+4 | 10 | 8 | 1(1), 1(2) | 0 |
| 1+2+5 | 5 | 4 | 0 | 1(4) |
| 1+4 | 5 | 3 | 1(1+2+4) | 1(5) |
| 1+6 | 5 | 4 | 1(1+6+7) | 0 |
| 4+5 | 10 | 8 | 1(4) | 1(2) |
| 6+7 | 15 | 12 | 1(1+6+7), 1(7) | 1(1) |
| Total | 120 | 95 | 17 | 8 |
| R | | 79.2% | | |
| F | | 16.7% | | |

| AUs | Sequences | Appearance (MDA [8])+ Geometric (2DBDA) | | |
|---|---|---|---|---|
| | | True | Missing or Extra | False |
| 1 | 10 | 5 | 2(1+2+4), 1(1+2) | 2(2) |
| 2 | 5 | 3 | 1(1+2+4) | 1(1) |
| 4 | 10 | 8 | 1(1+2+4), 1(1+4) | 0 |
| 5 | 10 | 8 | 1(4+5) | 1(5) |
| 6 | 10 | 8 | 1(1+6) | 1(7) |
| 7 | 5 | 4 | 0 | 1(6) |
| 1+2 | 20 | 17 | 1(2), 1(1+2+4) | 1(4) |
| 1+2+4 | 10 | 8 | 1(1), 1(2) | 0 |
| 1+2+5 | 5 | 4 | 0 | 1(4) |
| 1+4 | 5 | 3 | 1(1+2+4) | 1(5) |
| 1+6 | 5 | 4 | 1(1+6+7) | 0 |
| 4+5 | 10 | 8 | 1(4) | 1(2) |
| 6+7 | 15 | 11 | 1(1+6+7), 2(7) | 1(1) |
| Total | 120 | 91 | 18 | 11 |
| R | | 75.8% | | |
| F | | 19.2% | | |

| AUs | Sequences | Only Geometric (2DBDA) | | |
|---|---|---|---|---|
| | | True | Missing or Extra | False |
| 1 | 10 | 5 | 2(1+2+4), 1(1+2) | 2(2) |
| 2 | 5 | 3 | 1(1+2+4) | 1(1) |
| 4 | 10 | 8 | 1(1+2+4), 1(1+4) | 0 |
| 5 | 10 | 8 | 1(4+5) | 1(5) |
| 6 | 10 | 7 | 2(1+6) | 1(7) |
| 7 | 5 | 2 | 2(6+7) | 1(6) |
| 1+2 | 20 | 17 | 1(2), 2(1+2+4) | 0 |
| 1+2+4 | 10 | 7 | 1(1), 2(2) | 0 |
| 1+2+5 | 5 | 3 | 1(1+2) | 1(4) |
| 1+4 | 5 | 3 | 1(1+2+4) | 1(5) |
| 1+6 | 5 | 3 | 1(1+6+7) | 1(7) |
| 4+5 | 10 | 8 | 1(4) | 1(2) |
| 6+7 | 15 | 10 | 2(1+6+7), 2(7) | 1(1) |
| Total | 120 | 84 | 25 | 11 |
| R | | 70.0% | | |
| F | | 23.3% | | |

**Table 2.** Lower facial action unit recognition results (R= recognition rate, F= false alarm)

| Appearance (MBDA)+ Geometric (2DBDA) | | | | |
|---|---|---|---|---|
| AUs | Sequences | Recognized AUs | | |
| | | True | Missing or extra | False |
| 9 | 4 | 4 | 0 | 0 |
| 10 | 6 | 6 | 0 | 0 |
| 12 | 6 | 6 | 0 | 0 |
| 15 | 4 | 4 | 0 | 0 |
| 17 | 8 | 8 | 0 | 0 |
| 20 | 6 | 6 | 0 | 0 |
| 25 | 24 | 24 | 0 | 0 |
| 26 | 12 | 10 | 2(25+26) | 0 |
| 27 | 12 | 12 | 0 | 0 |
| 9+17 | 12 | 12 | 0 | 0 |
| 9+17+23+24 | 2 | 2 | 0 | 0 |
| 9+25 | 2 | 2 | 0 | 0 |
| 10+17 | 4 | 3 | 1(17) | 0 |
| 10+15+17 | 2 | 1 | 1(15+17) | 0 |
| 10+25 | 4 | 4 | 0 | 0 |
| 12+25 | 8 | 8 | 0 | 0 |
| 12+26 | 4 | 3 | 1(12+25) | 0 |
| 15+17 | 8 | 8 | 0 | 0 |
| 17+23+24 | 4 | 4 | 0 | 0 |
| 20+25 | 8 | 8 | 0 | 0 |
| Total | 140 | 135 | 5 | 0 |
| R | 96.4% | | | |
| F | 2.1% | | | |

| Appearance (2DBDA [7]+BDA [6])+ Geometric (2DBDA) | | | | |
|---|---|---|---|---|
| AUs | Sequences | Recognized AUs | | |
| | | True | Missing or extra | False |
| 9 | 4 | 4 | 0 | 0 |
| 10 | 6 | 5 | 0 | 1(17) |
| 12 | 6 | 6 | 0 | 0 |
| 15 | 4 | 4 | 0 | 0 |
| 17 | 8 | 8 | 0 | 0 |
| 20 | 6 | 6 | 0 | 0 |
| 25 | 24 | 20 | 2(25+26) | 2(26) |
| 26 | 12 | 10 | 1(25+26) | 1(25) |
| 27 | 12 | 12 | 0 | 0 |
| 9+17 | 12 | 12 | 0 | 0 |
| 9+17+23+24 | 2 | 2 | 0 | 0 |
| 9+25 | 2 | 2 | 0 | 0 |
| 10+17 | 4 | 0 | 3(10+12) | 1(12) |
| 10+15+17 | 2 | 1 | 1(15+17) | 0 |
| 10+25 | 4 | 4 | 0 | 0 |
| 12+25 | 8 | 8 | 0 | 0 |
| 12+26 | 4 | 2 | 2(12+25) | 0 |
| 15+17 | 8 | 8 | 0 | 0 |
| 17+23+24 | 4 | 4 | 0 | 0 |
| 20+25 | 8 | 4 | 4(20+26) | 0 |
| Total | 140 | 122 | 13 | 5 |
| R | 87.1% | | | |
| F | 12.1% | | | |

| Appearance (MDA [8])+ Geometric (2DBDA) | | | | |
|---|---|---|---|---|
| AUs | Sequences | Recognized AUs | | |
| | | True | Missing or extra | False |
| 9 | 4 | 4 | 0 | 0 |
| 10 | 6 | 3 | 2(10+7) | 1(17) |
| 12 | 6 | 6 | 0 | 0 |
| 15 | 4 | 4 | 0 | 0 |
| 17 | 8 | 7 | 1(10+17) | 0 |
| 20 | 6 | 6 | 0 | 0 |
| 25 | 24 | 21 | 2(25+26) | 1(26) |
| 26 | 12 | 9 | 3(25+26) | 0 |
| 27 | 12 | 12 | 0 | 0 |
| 9+17 | 12 | 12 | 0 | 0 |
| 9+17+23+24 | 2 | 0 | 2(9+17+24) | 0 |
| 9+25 | 2 | 2 | 0 | 0 |
| 10+17 | 4 | 1 | 2(10+12) | 1(12) |
| 10+15+17 | 2 | 1 | 1(15+17) | 0 |
| 10+25 | 4 | 4 | 0 | 0 |
| 12+25 | 8 | 8 | 0 | 0 |
| 12+26 | 4 | 0 | 3(12+25) | 1(25) |
| 15+17 | 8 | 8 | 0 | 0 |
| 17+23+24 | 4 | 3 | 1(17+24) | 0 |
| 20+25 | 8 | 5 | 3(20+26) | 0 |
| Total | 140 | 116 | 20 | 4 |
| R | 82.9% | | | |
| F | 14.3% | | | |

| Only Geometric (2DBDA) | | | | |
|---|---|---|---|---|
| AUs | Sequences | Recognized AUs | | |
| | | True | Missing or extra | False |
| 9 | 4 | 4 | 0 | 0 |
| 10 | 6 | 2 | 2(10+7) | 2(17) |
| 12 | 6 | 6 | 0 | 0 |
| 15 | 4 | 4 | 0 | 0 |
| 17 | 8 | 6 | 1(10+17) | 1(10) |
| 20 | 6 | 6 | 0 | 0 |
| 25 | 24 | 21 | 2(25+26) | 1(26) |
| 26 | 12 | 8 | 2(25+26) | 2(25) |
| 27 | 12 | 12 | 0 | 0 |
| 9+17 | 12 | 12 | 0 | 0 |
| 9+17+23+24 | 2 | 0 | 2(9+17+24) | 0 |
| 9+25 | 2 | 2 | 0 | 0 |
| 10+17 | 4 | 1 | 2(10+12) | 1(12) |
| 10+15+17 | 2 | 1 | 1(15+17) | 0 |
| 10+25 | 4 | 4 | 0 | 0 |
| 12+25 | 8 | 8 | 0 | 0 |
| 12+26 | 4 | 0 | 2(12+25) | 2(25) |
| 15+17 | 8 | 8 | 0 | 0 |
| 17+23+24 | 4 | 3 | 1(17+24) | 0 |
| 20+25 | 8 | 4 | 3(20+26) | 1(26) |
| Total | 140 | 112 | 18 | 10 |
| R | 80.0% | | | |
| F | 17.1% | | | |